\documentclass[letterpaper, 10 pt, conference]{ieeeconf}  

\IEEEoverridecommandlockouts                              

\usepackage{amsmath} 
\usepackage{amssymb}  
\usepackage{cite}
\usepackage{float}
\usepackage{graphicx}
\usepackage{textcomp}
\usepackage{xcolor}
\usepackage{booktabs}
\usepackage{algorithm}
\usepackage{algpseudocode}
\usepackage{stfloats}
\usepackage{balance}

\title{\LARGE \bf
Imitation and Adaptation Based on Consistency: A Quadruped Robot Imitates Animals from Videos Using Deep Reinforcement Learning
}

\author{Qingfeng Yao$^{1,2,3*}$, Jilong Wang$^{4*}$, Shuyu Yang$^{4}$, Cong Wang$^{1,2,3}$, Hongyin Zhang$^{4}$, \\Qifeng Zhang$^{1,3}$, Donglin Wang$^{4+}$
\thanks{* Contributed equally}
\thanks{+ Corresponding author. Email: wangdonglin@westlake.edu.cn}%
\thanks{The main work was done at MiLAB, Westlake University.}
\thanks{$^{1}$State Key Laboratory of Robotics, Shenyang Institute of Automation, Chinese Academy of Sciences, Shenyang 110016, China}%
\thanks{$^{2}$Institutes for Robotics and Intelligent Manufacturing, Chinese Academy of Sciences, Shenyang 110169, China}%
\thanks{$^{3}$University of Chinese Academy of Sciences, Beijing 100049, China}%
\thanks{$^{4}$School of Engineering, Westlake University, Hangzhou 310024, China}%
}

\begin{document}

\maketitle
\thispagestyle{empty}
\pagestyle{empty}

\begin{abstract}

The essence of quadrupeds' movements is the movement of the center of gravity, which has a pattern in the action of quadrupeds. However, the gait motion planning of the quadruped robot is time-consuming. Animals in nature can provide a large amount of gait information for robots to learn and imitate. Common methods learn animal posture with a motion capture system or numerous motion data points. In this paper, we propose a video imitation adaptation network (VIAN) that can imitate the action of animals and adapt it to the robot from a few seconds of video. The deep learning model extracts key points during animal motion from videos.
The VIAN eliminates noise and extracts key information of motion with a motion adaptor, and then applies the extracted movements function as the motion pattern into deep reinforcement learning (DRL). 
To ensure similarity between the learning result and the animal motion in the video, we introduce rewards that are based on the consistency of the motion. DRL explores and learns to maintain balance from movement patterns from videos, imitates the action of animals, and eventually, allows the model to learn the gait or skills from short motion videos of different animals and to transfer the motion pattern to the real robot.  
\normalcolor

\end{abstract}

\section{INTRODUCTION}

Currently, three major types of robots are well developed in locomotion:  wheeled, tracked, and legged robots. Compared to robots in other categories, legged robots have shown significantly higher performance when facing rough terrains due to their unique ability to have discrete ground contact points through their gait \cite{lee2020learning}. They can adjust their foothold position to alter their gaits according to the characteristics of terrain and tasks. However, legged robots must consider a massive number of factors to search for an optimum in high dimensional action space \cite{grandia2020multi}. 

\begin{figure}[tbp]
\centering
\includegraphics[width=0.4\textwidth]{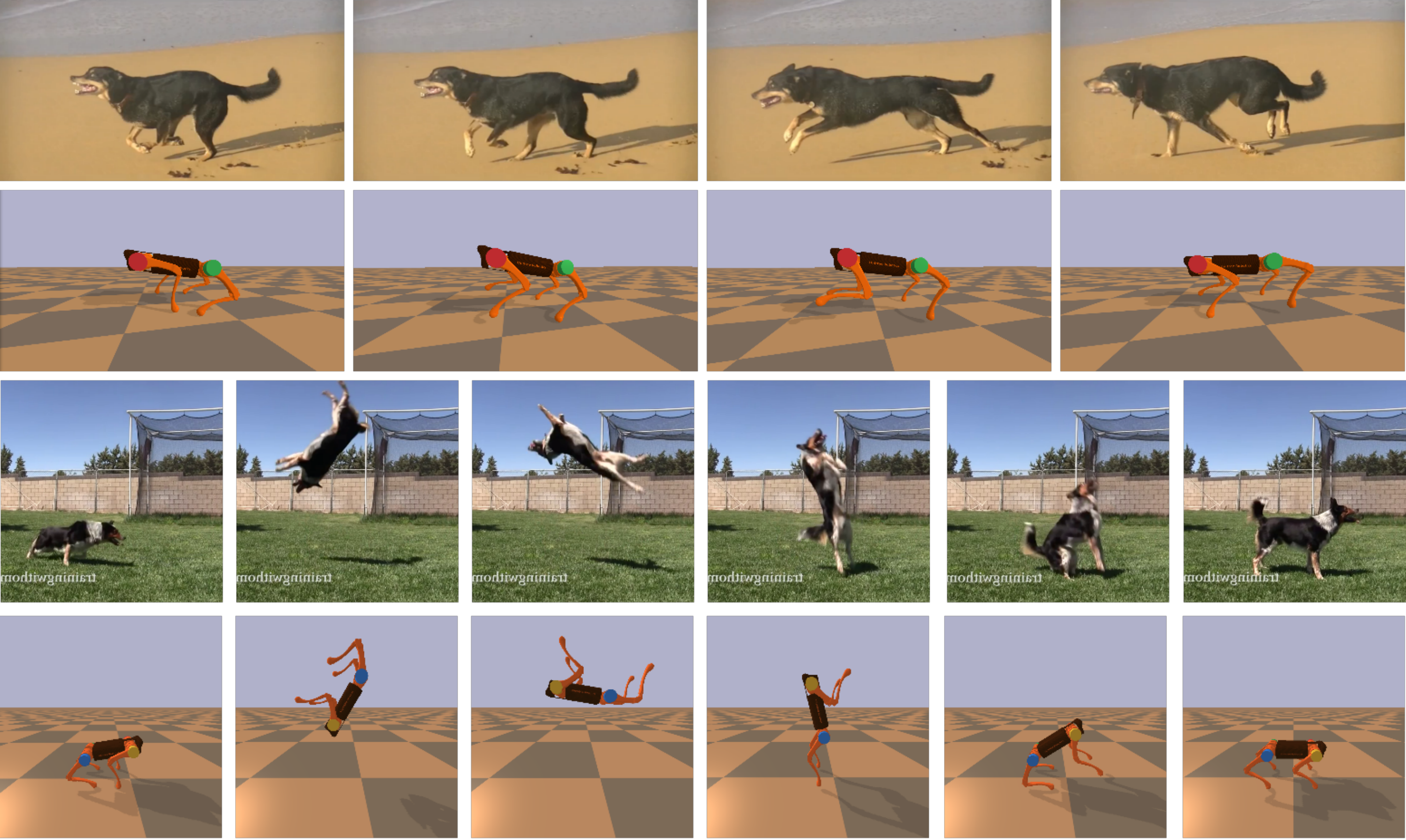}
\caption{
Quadrupedal robots can learn different skills from few-second videos (the first and third rows), such as galloping (the second row) and backflip (the fourth row). 
}
\label{Fig.Intro}
\end{figure}

When straddling, it is critical to that the robot's center of gravity projection and zero moment point remain within its geometry to maintain movement stability\cite{winkler2017fast}. Traditional algorithms often plan the center of gravity trajectory in the form of discrete cycles and reduce the speed or increase the frequency to ensure stability. Nature is the teacher of humans, and many designs of robots are inspired by animals. To allow the robot to learn the action of animals more efficiently, it is reasonable to apply imitation learning to robots. Imitation learning helps robots imitate the action of animals from existing animal data and transfer this knowledge to identical tasks, and the demonstration of animals is regarded as fundamental knowledge.

At the same time, deep reinforcement learning (DRL) has emerged rapidly from machine learning in recent years, and has gradually shown its potential in decision-making problems. 
In this paper, we propose a network that combines deep learning and DRL for quadruped robots to learn cyclical and noncyclical movements through animal videos. The imitation learning framework includes recognition, imitation and adaptation. 
Our framework combines the benefits of DRL and deep learning, allowing quadruped robots to learn various behavior patterns using only a few seconds of animal videos. DRL control networks can investigate and learn how to adapt information from various animals to complete tasks by adjusting their posture and maintaining motion stability. The framework proposed in this paper is trained in a simulation environment and transferred directly to the real quadruped robot. 
In summary, this work made the following contributions:

\begin{itemize}%
\item The key animal nodes are extracted from the video through the deep neural network as a reference trajectory and a motion adaptor is used to remove the offset and error from the key nodes. 
\item Features extracted by the motion adaptor are prior information of the DRL network. DRL explores how to maintain balance based on prior information from animal videos. 
\item Simulation and real experiment with different type of motion demonstrate the ability of our algorithm.
\end{itemize}

\section{Related Work}

Traditional legged robot control methods focus on modeling the robot and the surrounding environment, followed by positional estimation and trajectory planning using forward and inverse kinematics \cite{raibert1986legged}. Carlo et al. \cite{di2018dynamic} proposed a model-based method for the calculation of forces, which can be summarized as a convex optimization problem improving the effectiveness of MPC in quadruped robots.
Bellicoso et al. \cite{bellicoso2018dynamic} proposed an online motion planner based on the zero moment point. The position of the center of mass was optimized for the execution of different gaits.

In recent years, machine learning has played an increasingly important role in quadruped robot control.
Villarreal et al. \cite{villarreal2019mpc} combined a convolutional neural network with MPC. Yang et al. \cite{yang2020multi} used a gating neural network combined with pretrained neural networks to adapt to changes in the environment and tested the combined method on a real robot.

DRL is a learning process that enables agents to learn from scratch by feedback from exploring the environment\cite{mnih2015human}.
Bellegarda et al. \cite{bellegarda2020robust} proposed a framework combining DRL with nonlinear trajectory optimization to assist quadruped robots in jumping.
Lee et al. \cite{lee2020learning} proposed a robust DRL-based controller resulted from curriculum training to guide a quadruped robot to locomote on complex terrains without any visual aid.
Escontrela et al. \cite{escontrela2020zero} proposed a method using simulated 3D lidar sensors to complete navigation tasks on different terrains with a unified policy.

Imitation learning can swiftly master new tasks by observing expert demonstrations and achieving a feasible reproducibility of the expert strategy\cite{hussein2017imitation}. A strategy that directly maps trajectory actions by state features with a method called behavior cloning (BC)\cite{pomerleau1991efficient} was developed to directly learns the state-to-action mapping. BC transforms imitation learning into a supervised learning problem, which is easily affected by changes in state distribution, therefore, a regularized behavioral cloning imitation learning method was proposed\cite{reddy2019sqil}. In this method, the soft-Q-learning algorithm was introduced to overcome the disturbance from the imbalance of state distributions.
Another task is to convert the imitation learning problem into a DRL problem by inferring the reward function from the expert demonstration\cite{ho2016generative}.

Imitation learning already has applications in robots. Xie et al. \cite{xie2019iterative} proposed a deterministic action stochastic state method that uses the policy gradient to help the biped robot Cassie generate gaits that are similar to those of humans. A method that learns basketball dribbling control from motion capture data was proposed by Liu \cite{liu2018learning}. The system developed a strategy based on trajectory optimization and DRL to learn the skill of dribbling. A controller was developed by Yamane \cite{yamane2010controlling} to allow the robot to maintain balance while tracking the movement of a given reference. Peng et al. \cite{peng2018deepmimic} proposed a method that could handle key-frame motion. By combining imitation targets with task goals, the agents were trained to react intelligently to the environment and perform a variety of skills.

3D pose estimation is a way to take full advantage of animal demonstrations with multiple cameras \cite{nath2019using}.
A motion-sensing camera is used to extract the demonstration information, and the Levenberg-Marquardt method is used to optimize the solution of inverse kinematics to improve the stability and similarity of the robot movement from human posture \cite{hu2014online}. 

A framework based on pose estimation and DRL was used to extract the whole body 3D reference motion from public video clips \cite{peng2018sfv}.
Kearney et al. \cite{kearney2020rgbd} studied the 3D pose estimation problem of dogs based on RGBD images by using a single motion capture system to obtain the skeleton. Ou et al. \cite{ou2015real} addressed this issue by planning the zero moment point and driving angle through human body demonstration data so that the robot could imitate human actions and gait patterns.

The pose skills of animals can be used by quadruped robots because of their similar morphologies.
An imitation learning system was proposed by Peng \cite{peng2020learning} that enables quadruped robots to learn agile motion skills by imitating animals, where mocap 3D data is used to match and retarget various behaviors for imitation learning. 
Capturing motion data with a 3D camera were a requirement for the type of equipment and data\cite{holden2016deep}. 

Monocular camera videos are fairly easy to obtain; therefore Vondrak et al. \cite{vondrak2012video} took a more economical approach by estimating human motion patterns from monocular videos. A state-space biped controller with a balanced feedback mechanism executed the required movement by directly establishing it from the monocular video.
Da et al. \cite{da2020learning1} proposed a method to imitate different sports, where the retargeting method was applied to the motion capture data, and the inverse kinematics algorithm was used for imitation. The generation of a motion trajectory was transformed into a nonlinear optimization problem to minimize the trajectory error between the animal and the robot.

Learning from animal video technology provides a simple and effective method for animal behavior imitation, but manually extracting key nodes is time-consuming. To solve this problem, an unlabeled pose estimation method \cite{mathis2018deeplabcut} based on deep neural network transfer learning is proposed to track different key body nodes of different species with minimal training data.

\begin{figure}[tbp]
\centering
\includegraphics[width=0.4\textwidth]{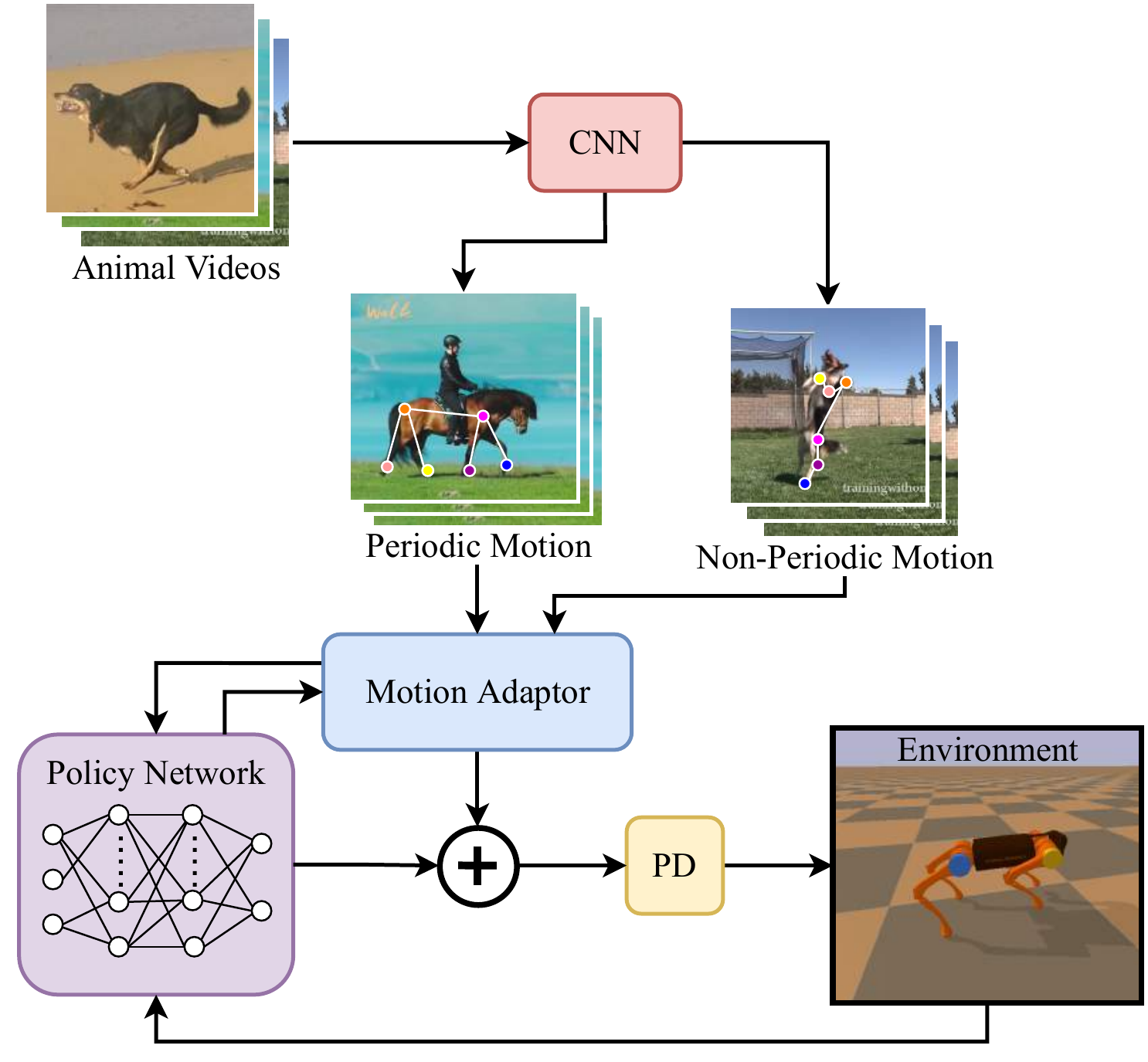}
\caption{
The overview of VIAN. The imitation learning framework includes the part of recognition and imitation based on deep learning and DRL. 
}
\label{Fig.overview}
\end{figure}

\section{Method}
We propose a novel \textit{video imitation adaptive network (VIAN)} framework based on animal videos to imitate learning and adapt to the robot by mapping the states to the corresponding actions with neural networks.

As shown in Fig. \ref{Fig.overview}, first, we analyzed the motion of the animal and extracted key nodes with the deep neural network. To compensate for the structural difference between the animal and the quadruped robot, we selected key points on the body of the animal that could be mapped to the corresponding body structure of the quadruped robot. For example, the paw of a dog matched the foot of the robot, and the neck of a dog matched the front of the body of the robot. 
We designed a motion adaptor that filtered out noise and offsets from key nodes to formulate a reference trajectory for the robot to imitate.
Finally, we used DRL to train the robot to imitate the motion and maintain balance with a consistency reward.

Here we employ a key frame extraction method based on \textit{DeepLabCut} \cite{mathis2018deeplabcut} to obtain the demonstration features from the videos.
DeepLabCut is a toolbox used for unmarked pose estimation of animals performing various tasks. We tested animals such as a running dog or a pacing horse. 
In the movement of a quadruped robot, the position of feet and legs is the control variable. Therefore, we extracted the positions of the head, buttocks and feet from moving animals as the initial information. 
\subsection{Motion Adaptor}

To allow the robot to complete different tasks by imitating animals in the videos, we separate the animal motion in the videos into two types: periodic motion and aperiodic motion. 
To map the body of the animals onto the structure of the robot, we selected a set of key points that were specified on the animal and that were paired with corresponding target key points on the robot.

\begin{figure}[htbp]
\centering  
    \includegraphics[width=0.4\textwidth]{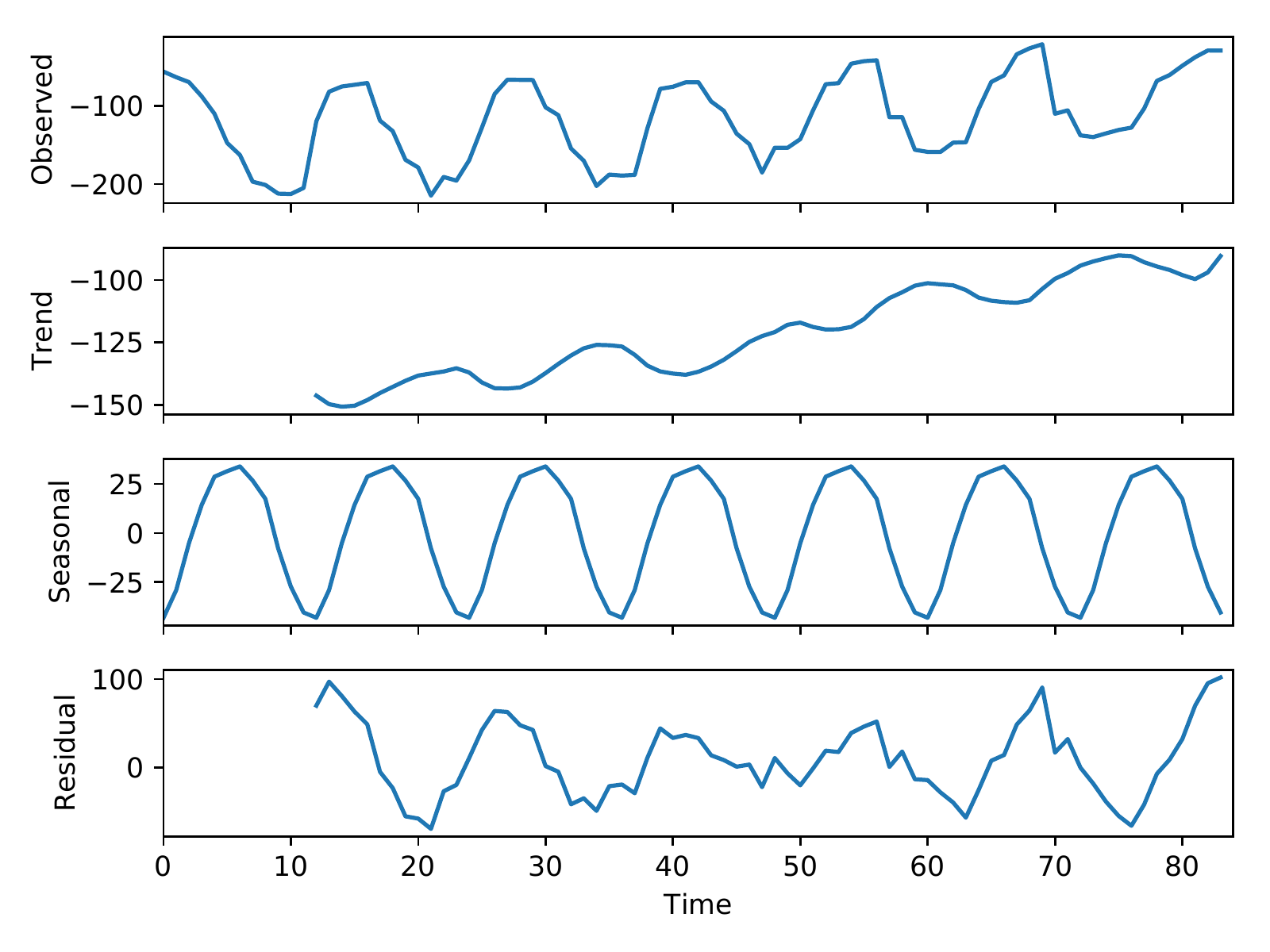}
\caption{
Original message divided into the trend, seasonal and residual factors by X11.
}
\label{Fig.3}
\end{figure}

Moving limbs tend to deflect and cause additional noise due to angular variations in periodic motion video.
To avoid video noise and angle deviation in the moving process, we separated the movements in the animal video into three sections for periodic motion such as walking and running.
The time series data method X11 \cite{dagum2016seasonal} was used to separate the data into long-term trends, seasonal trends and random components. The X11 decomposition method is based on the classic time series decomposition method, and it overcomes the shortcomings of the classic time series decomposition method. Specifically, it can estimate the trend-period term of each period, including the endpoints, and it allows seasonal slow changes. The model uses iterative process fitting to estimate three X11 factors: trend cycle, seasonality, and noise. The X11 method assumes that the main components of the time series follow the additive model: 
\begin{equation}
y_{t}=\mathrm{TC}_{\mathrm{t}}+S_{t}+I_{t},
\end{equation}
where $y_t$ represents the original series, $TC_t$ represents the trend, $S_t$ represents the periodicity, and $I_t$ represents the residual. The components were estimated using a smoothing linear filter. The filters were applied in order, and finally, a periodic function was obtained as a function of the seasonality of motion. 
The X11 method was used to process the movement of the x-axis and z-axis of key nodes obtained in video over time.
Fig. \ref{Fig.3} shows the trajectory of motion and the effect after processing with the X11 method. The X11 method removes the noise and trends in motion and extracts the periodic function $S_{t}$ as the imitation information. The period is the X11 input parameter, which is calculated from the average peak distance of animal videos.


\begin{figure}[tbp]
\centering
    \includegraphics[width=0.45\textwidth]{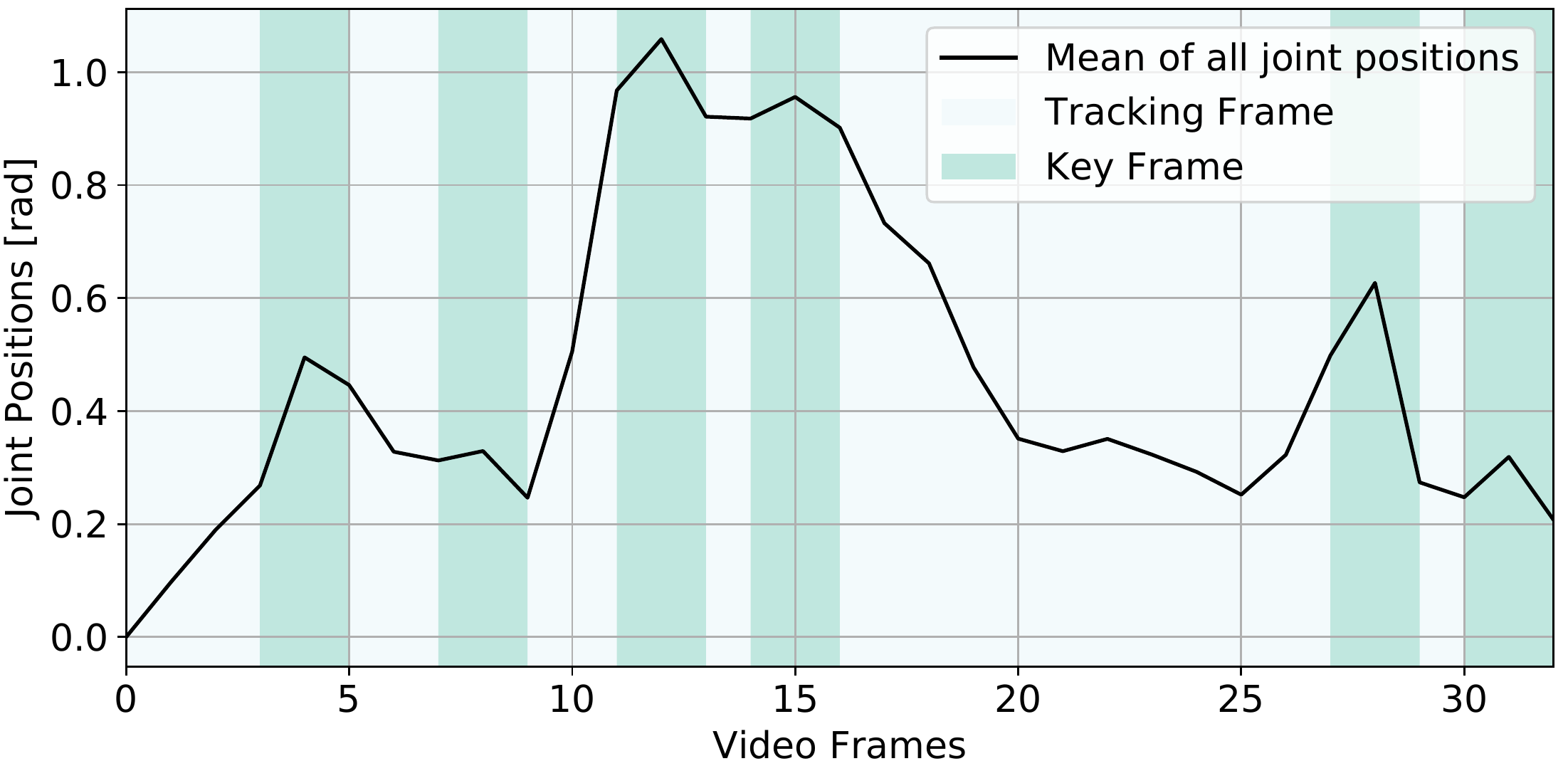}
    \caption{
    The key frame is extracted according to foot movement
    }
\label{Fig.key_frame}
\end{figure}

Due to the different body proportions of the animals in the videos, we preliminarily process key points of the video according to the body proportions of the quadruped robot and the extracted key node positions.
The scale of the zoom is:
\begin{equation}
f_{i}(t)=\frac{X^{lim}_i}{Max(S_i)} * S_{i}(t).
\end{equation}
The maximum range that the \textit{i-th} leg can move is $X^{lim}_i$ and $Max(S_i)$ is the range of the \textit{i-th} leg periodic function $S_i(t)$. To avoid the inaccuracy and unsteady motion behavior induced by applying $f_{i}(t)$ directly to a quadruped robot and performing the movement, the quadruped robot needs to adapt video motion information to its own movement process and learn to adjust $f_{i}(t)$ according to its current state.

By combining posture imitation and DRL, the robot learns to adapt to the periodic function according to its posture and completes different movements, after which the DRL network adjusts the periodic function $f_i(t)$. 
DRL outputs are determined by the internal state, which includes three historic motor positions, the IMU, and the foot position.
We calculate the final quadruped position as:
\begin{equation}
g_{i}(t)=A_{i}(t)\left(f_{i}(t)\right)+b_{i}(t),
\end{equation}
where $A_i(t)$ is the prior feature scaling and $b_i(t)$ is the adjustment offset output by the neural network. Finally, the angles of the leg joints are obtained by solving the inverse kinematics with $g_i(t)$. 
At the same time, the network outputs $\Delta T$ to control the motion frequency of the robot.

For aperiodic motion, since there are no seasonal data for the robot to learn the repeated motion cycle, we marked the key frames of actions that had the most impact on finishing the complete movement. For example, in the backflip task, actions such as jumping and pitching back are essential for the rotation of the body. To determine the key frames, we determined that the average angle of the video joint yields the key frame. It becomes a key frame when the average angle $\theta(t)$ reaches the local maximum value with threshold $\delta_{angle}$:
\begin{equation}
\theta_{mean}(t)-\theta_{mean}(t-1)>\theta_{mean}(t+1)-\theta_{mean}(t)+\delta_{angle}
\label{'beta'}
\end{equation}

The robot is rewarded for walking forward and maintaining a consistent body direction. 
The episode terminates when the robot loses its balance or walks out of a designated area. The reward function for this task is written as fellow: 

\begin{equation}
r_{lin}:=\left\{\begin{array}{ll}
\exp \left(-5.0\left(v_{x}-0.3\right)^{2}\right) & v_{x}<0.3 \\
1 & v_{x}>=0.3 \\
\end{array}\right.,
\end{equation}

\begin{equation}
r_{ang}:=\exp \left(-3.0\left(\omega_{y}\right)^{2}\right) ,
\end{equation}

and 
\begin{equation}
r_{body}:=\exp \left(-3 v_{y}^{2}\right)+\exp \left(-3.0\left(\theta_{r} +\theta_{y}\right)^{2}\right),
\end{equation}
where $r_{lin}$ indicates that the robot moves forward and $r_{ang}$ and $r_{body}$ limit the linear velocity tangent to the target direction and the attitude angle, and ensure that the robot does not deviate from the forward direction. 
$v_x$ and $v_y$ represent linear velocities along the x-axis and y-axis, respectively
and $\theta_{r}$, $\theta_{y}$ and $\omega_{y}$ represent roll, yaw and the yaw velocity, respectively.

\subsection{Consistency Reward}
When analyzing the animal walking data, we only focus on one set of data that contains time steps and the Cartesian coordinates of the corresponding key points. To better adapt to the motion, we introduced gait rewards to strengthen the ability of the robot to imitate the capture video actions. They are divided into two parts. One reward type is the stamp consistency reward that keeps the robot landing in sync with the video animal. One key parameter in quadruped walking is the ratio of the leg swinging phase to the entire cycle. According to the range of landing, the robot learns how to synchronize its feet to keep track of the animal. Another reward is the motion consistency reward. It is difficult to directly capture the speed and angular velocity in the 2D video, so we use the direction of the relative speed for their determination. This reward helps the robot move in the same direction as the animal, and to have the same orientation of momentum.

\begin{figure}[htbp]
\centering

    \includegraphics[width=0.45\textwidth]{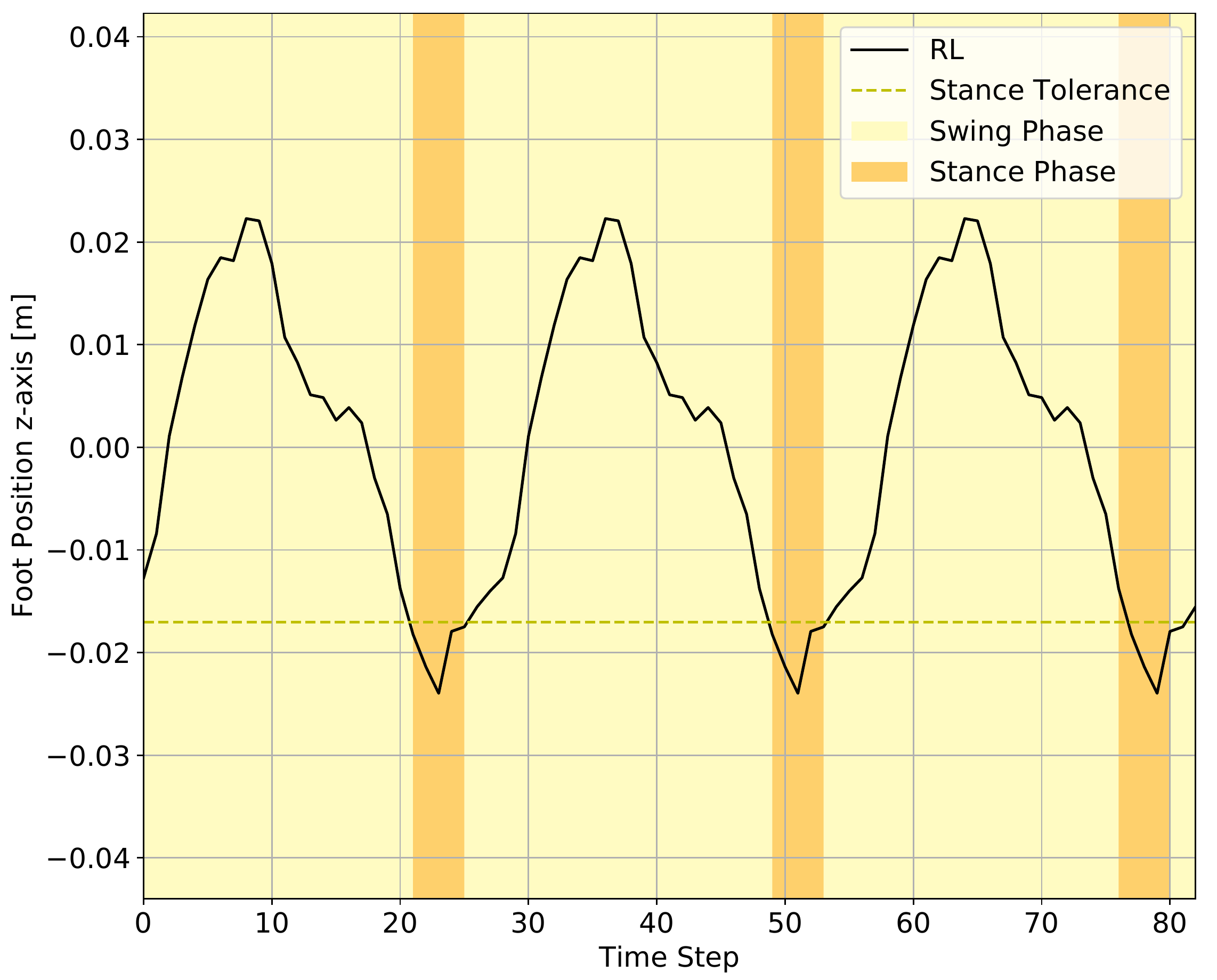}
    \caption{
    Foothold location and foot lifting area of the animal in the video obtained by calculating the swing leg thresholds. The consistency of the gaits is proven if the foot of the robot is on the ground.
    }
\label{Fig.foot_judge}
\end{figure}

We calculate the changes in foot height by analyzing the motion trajectory of the foot in the video. The trajectory threshold line is determined from $(h_i^{max}-h_i^{min})*\gamma$, where $\gamma$ is the threshold coefficient and $h_i^{max}$ and $h_i^{min}$ are the maximum and minimum of the foot height for different feet $i$ in $f_i(t)$. When the foot motion trajectory is lower than the threshold line, the animal foot position is close to the ground; hence, the robot foot is on the ground in the simulation. Punishment is applied when the gait trajectories appear dissimilar, as shown in Fig. \ref{Fig.foot_judge}.

The curve of the foot swing z-axis value of the animal changing with time is shown in Fig. \ref{Fig.foot_judge}, where we set threshold $\gamma$ to 0.1, as shown by the dotted line. When the curve is above the threshold, the \textit{i-th} foot is considered to be in the air and an indicator of the video foot position, with $d_i=0$; when the curve is beneath the threshold, the foot is considered to be on the ground and $d_i=1$.

$p_i$ is the foot position indicator for the robot. The \textit{i-th} foot $p_i$ is on the ground when $p_i=1$; otherwise, $p_i=0$. At the beginning of training, the robot did not learn to lift its leg. Hence, to encourage its robot to learn to lift the leg in the same manner as the animal in the video, we introduce the following stamp reward:

\begin{equation}
r_s^i=\left\{\begin{array}{ll}
-1 & \text { if } d_i=0 \cup p_i=1 \\
0  & \text { other }
\end{array}\right.
\end{equation}

We further propose motion consistency to enhance the imitation effect.
We estimate the accuracy of imitation by calculating the positions of the feet relative to the rear and front. To calculate the positions of feet relative to the body, the position angle of the head$(x^1,y^1)$ and rear$(x^2,y^2)$ is computed:
\begin{equation}
\beta=\arctan \frac{\left(y^{1}-y^{2}\right)} {\left(x^{1}-x^{2}\right)}.
\end{equation}

The feet positions are processed based on the relative angle of the front and rear. We obtain the \textit{i-th} foot’s motion state$(x_i,y_i)$ related to location $(x_{body}^i,y_{body}^i)$ based on the animal body frame in the video as follows:

\begin{equation}
\left\{\begin{array}{ll}
x_{body}^i=x_i \cos (\beta)+y_i \sin (\beta) \\
y_{body}^i=y_i \cos (\beta)-x_i \sin (\beta)
\end{array}\right.
\end{equation}

The consistency of the motion is evaluated by calculating the foot’s motion direction relative to the body frame. To encourage the foot of the robot to move in the same direction as the reference action, the motion consistency reward is calculated by comparing the computed moving speed of feet and the reference moving direction as shown below:

\begin{equation}
r_m^i=\left\{\begin{array}{ll}
0 & \text { if } \left\|v_{body}^i\right\|<\delta \\
\frac{v_i^{body}*v_i^{real}}{\left\|v_i^{body}*v_i^{real}\right\|} & \text { if }  \left\|v_{body}^i\right\|>=\delta
\end{array}\right .
\end{equation}
where $\delta$ is the speed threshold, $v_i^{body}$ is the \textit{i-th} foot speed of movement in the video calculated from $x_{body}^i$ and $y_{body}^i$, and $v_i^{real}$ is the speed of the \textit{i-th} foot of the robot.

\begin{figure}[tbp]
\centering

    \includegraphics[width=0.45\textwidth]{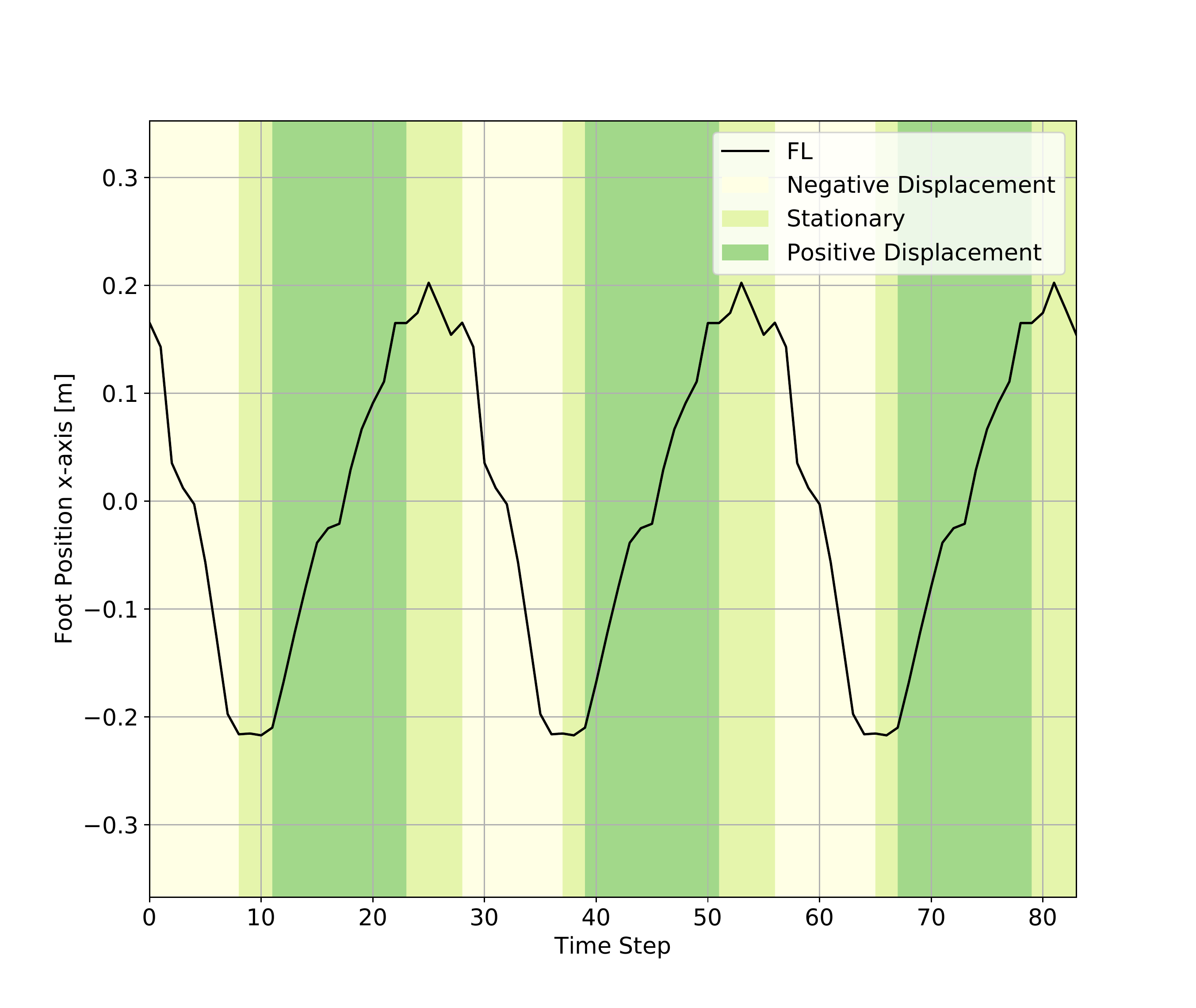}
    \caption{
    The front left foot position trending.
    Different colors present different statuses of motion: low speed is light green, moving forward is dark green, and backward is white.
    }
\label{Fig.trending}
\end{figure}

The foot motion of the animal along the x-axis is shown in Fig. \ref{Fig.trending}. We partition the moving process into 3 stages based on velocity, which is presented in different colors based on their values. 

The above rewards are summed into the total reward in periodic tasks:
\begin{equation}
\begin{aligned}
r_{total}=2*r_{lin}+2*r_{ang}+r_{body}+r_{f}+r_{g}
\end{aligned}    
\end{equation}

We further extended the angle consistency index to fix this situation in the backflip task. 
We refer to the angle of animal $\beta_{ref}$ in the video and return the corresponding reward according to the angle $\beta_{real}$ of the robot in the simulation as follows:

\begin{equation}
r_a = exp(-5(\beta _{ref}-\beta _{real})^2)
\end{equation}

Rewards are composed of angle consistency, stamp consistency and motion consistency in backflip tasks:
\begin{equation}
r_{total} = 2*r_a+r_m+r_s
\end{equation}

Reinforcement learning directly outputs the joint angle and time interval in this scenario.

\section{Experiment}
We first trained our policies in an open-source simulation, PyBullet, and then evaluated them on a \textit{Unitree} A1 robot, which is a quadruped robot with 12 degrees of freedom (3 per leg). 
Meanwhile, to implement the RL algorithm on real robots,
sim-to-real problems \cite{tan2018sim} still need to be solved to bridge the gap between simulation and reality.
Common solutions include domain adaptation\cite{bousmalis2018using} and randomization\cite{peng2018sim}.
We randomize the parameters in the simulation so that the algorithm can be more robust to measure inaccuracies and uncertainty in the real world.
Details of parameter randomization are listed in TABLE \ref{tab:physical_parameters}.
\begin{table}[htp]
 \centering
  \caption{Range of physical parameters.
  At the beginning of each training, the physical parameters are uniformly sampled within these ranges.
  }\label{tab:physical_parameters}
\begin{tabular}{lc}
\toprule
Physical parameters& Range\\
\midrule
Mass& [0.8, 1.2]*defaults\\
Inertia& [0.5, 1.5]*defaults\\
Motor strength& [0.8, 1.2]*defaults\\
Motor friction& [0, 0.05]$Nms/rad$\\
Latency& [0.0, 0.04]$s$\\
Lateral friction& [0.5, 1.25]$Ns/m$\\
Battery& [14.0, 16.8]$V$\\
Joint friction& [0, 0.05]$Nm$\\
CoM position noise& [-5,5]$cm$\\
External force&[-3,3]$N$\\
Step height & [0.02, 0.06] $m$\\ 
Step width & [0.18, 0.23] $m$\\
\bottomrule
\end{tabular}
\end{table}

\begin{figure}[htbp]
\centering  
    \includegraphics[width=0.45\textwidth]{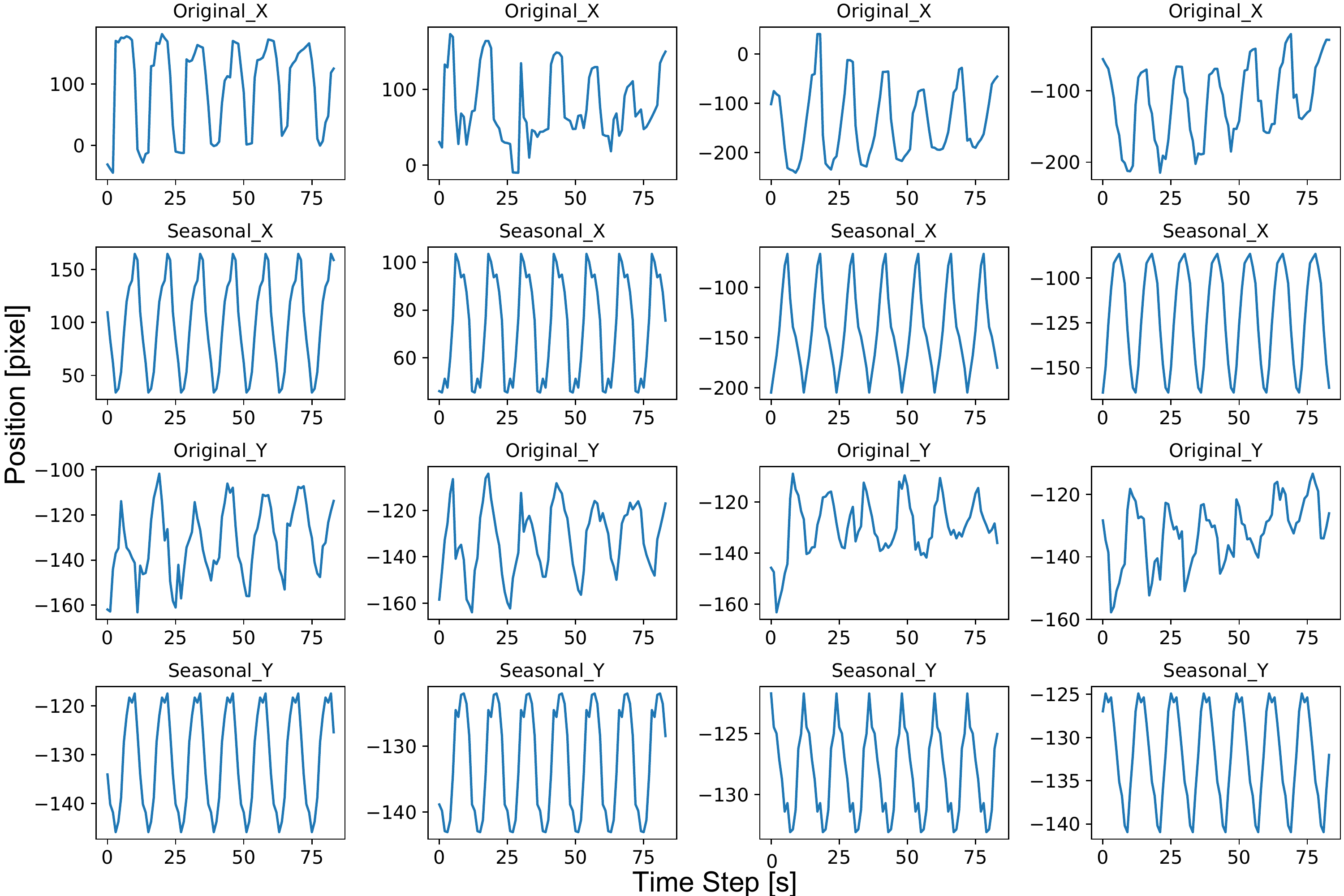}
\caption{
Effect of the motion trajectory of the dog’s feet extracted by X11. There is an observable improvement in periodicity and the elimination of noise.
}
\label{Fig.2}
\end{figure}

We extract video clips of walking and trotting from different animals, such as horses and dogs, with each clip being between only 3 seconds and 8 seconds long and containing several motion cycles. Then, the periodic function is obtained by the X11 process. We extract the x and z coordinates of the foot positions using X11 methods to extract seasonal motion, as shown in Fig. \ref{Fig.2}.
The result illustrates how the X11 process marks the positions of the dog’s feet during running and the increase in periodicity and stability.

To maintain the stability and consistency of motion, we scale seasonal motion to the range of the quadruped robot and obtain the basic motion trajectory, DRL is used to complete the movement with seasonal motion.
Adam is used as the training optimizer to train the dataset for the network with a learning rate of 0.001. The strategy network completes the convergence for each animal after 3e7 iterations of training. Robot training is performed with NVIDIA-2080 for 5 hours using Ubuntu 18.04.

Videos of moving horses and running dogs were used to test the effect of VIAN. 
We compared the effect of using DRL to train the data without X11 processing and applying X11 data directly to the robot.
The task was to move steadily forward for 10 meters. We conducted 20 experiments and recorded the success rate, corresponding speed and consistency reward. 
The results are shown in Table \ref{tab}. 
The quadruped robot cannot complete the task by directly using the extracted data.
Meanwhile, due to data noise, the result of training with data not processed by X11 has a lower effect.

    \begin{table}[t]
    \centering
    \caption{Effects of different tasks.
    }\label{tab}
    \begin{tabular}{lllll}
    \toprule
     Type             & Success Rate      & Speed      & Stamp      & Movement\\ \midrule
    RL(dog)              & 0.55 & 0.21 & 0.16 & 0.37  \\
    X11(dog)               & 0.0  & 0.05  & 0.2  & 0.52   \\
    VIAN(dog)              & 0.8 & 0.32 & 0.21 & 0.36   \\
    RL(horse)              & 0.55 & 0.42 & 0.06 & 0.41  \\
    X11(horse)              & 0.0  & -0.02  & 0.12  & 0.29   \\
    VIAN(horse)              & 0.75  & 0.48 & 0.08  & 0.42   \\
    \bottomrule
    \end{tabular}
    \end{table}

\begin{figure}[htbp]
\centering

    \includegraphics[width=0.5
    \textwidth]{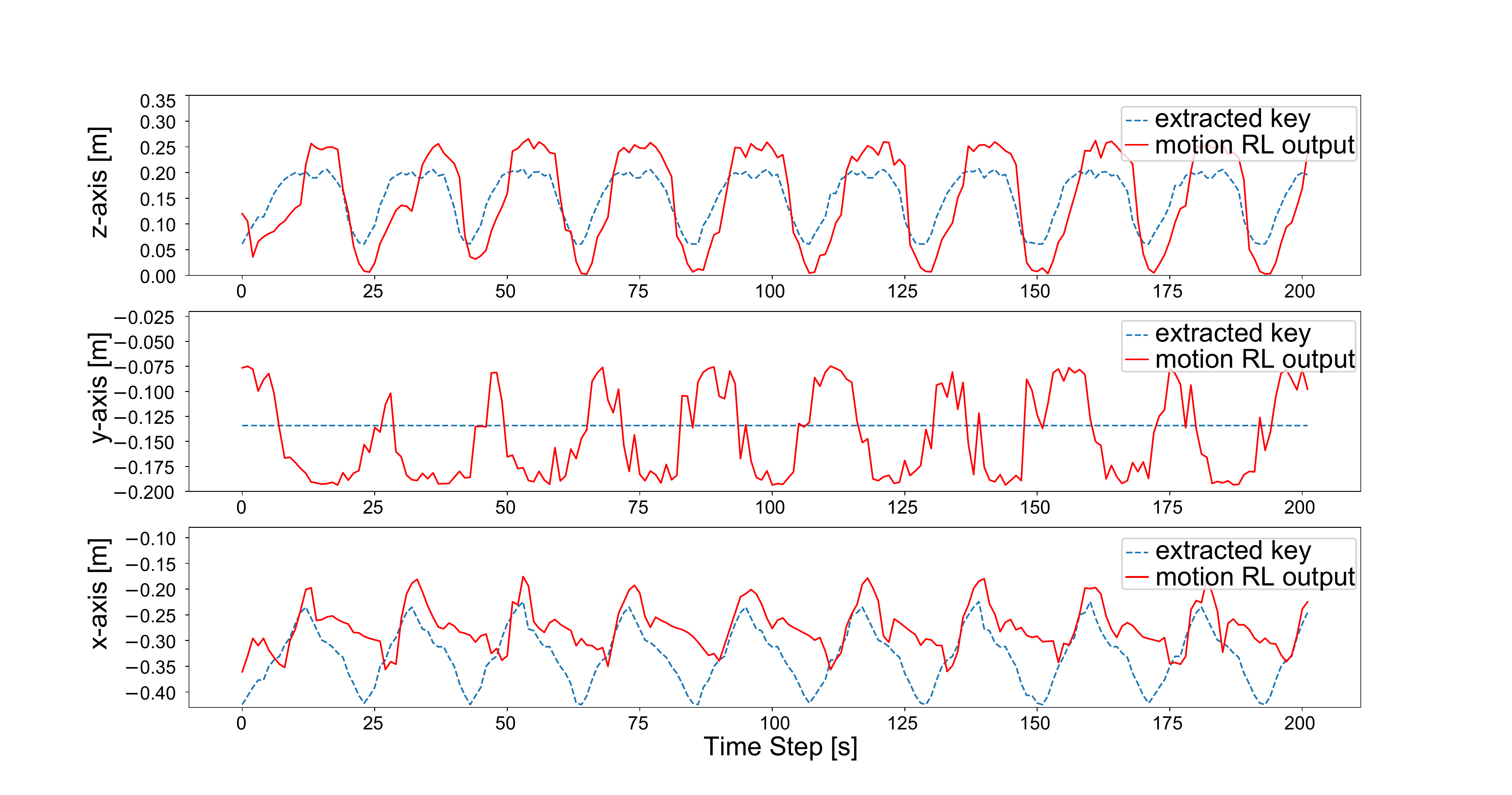}
    \caption{
    Result of DRL training. A periodic function was learned by DRL to help the robot maintain its gait according to the baseline knowledge.
    }
\label{Fig.run}
\end{figure}

To illustrate the effect of the model, we visualize the experimental results. The scaling curve of the limbs from the network when learning from a horse walking is shown in Fig. \ref{Fig.run}. The result shows the result of the robot imitating the horse walking. 
By visualizing the transformation of the motion trajectory, we observe that the robot can learn to maintain its body balance by adjusting the Y-axis even without information about the motion in the Y direction.

Meanwhile, RL provides a suitable scaling ratio and residual to transfer the original curve to a new curve that is suitable for the robot structure. The robot learns to scale up the moving distance along the x-axis, outputs a periodical residual to adjust the position along the y-axis, and smooths the original curve to help maintain the balance of its moving status on the z-axis. 

\begin{figure}[htbp]
\centering
    \includegraphics[width=0.45\textwidth]{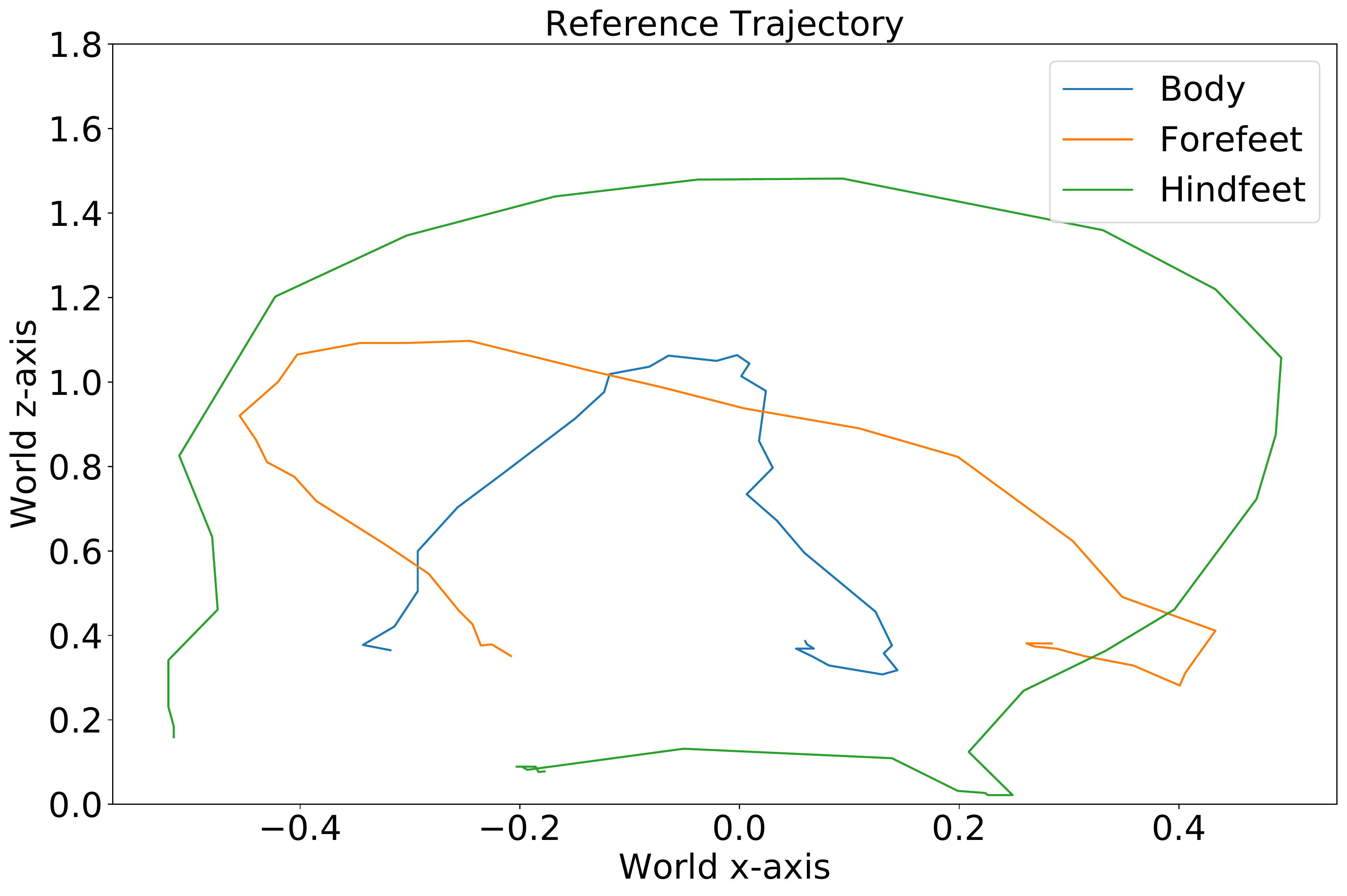}
    \includegraphics[width=0.45\textwidth]{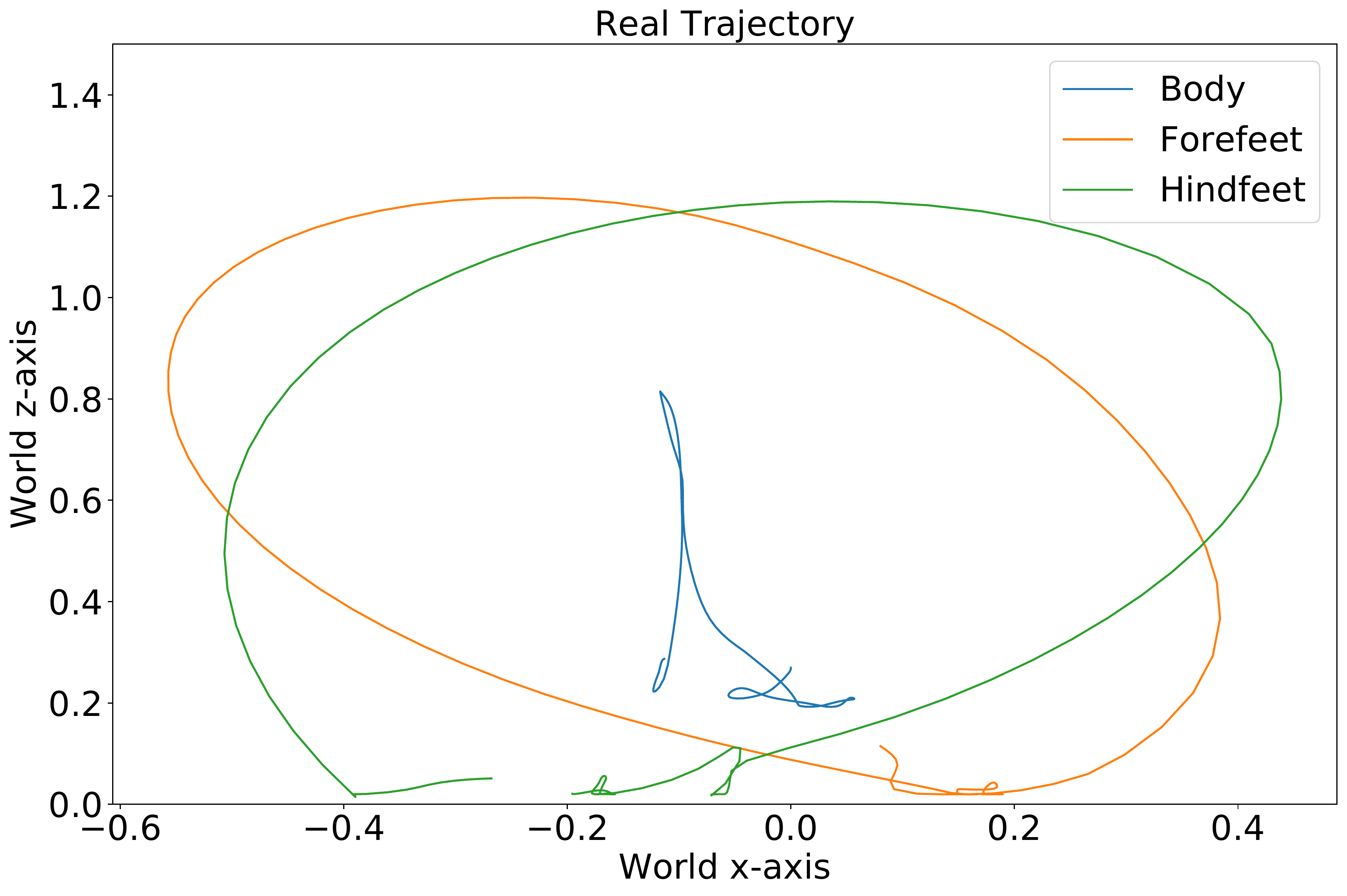}
    \caption{
    Result of DRL training Obtained by referring to the video and the final result of the x- and z-axis coordinates of the body center and front and back feet. Through key frames, the backflip skill is learned by DRL.
    }
\label{Fig.flip}
\end{figure}

To test the adaptability of our method, we attempted to implement the backflip motion on the quadruped robot. In this task, the key frames of the motion are extracted, and the DRL adjusts the joint angle of the corresponding key frame and the time interval of different key moments according to the reward. The final motion trajectory is shown in Fig. \ref{Fig.flip}. Combined with our framework, we used data from a few key frames of the video to complete the robot's backflip task. To the best of our knowledge, this is the first time an agile robot has completed the backflip task with DRL.

\begin{figure}[htbp]
\centering
    \includegraphics[width=0.11\textwidth,]{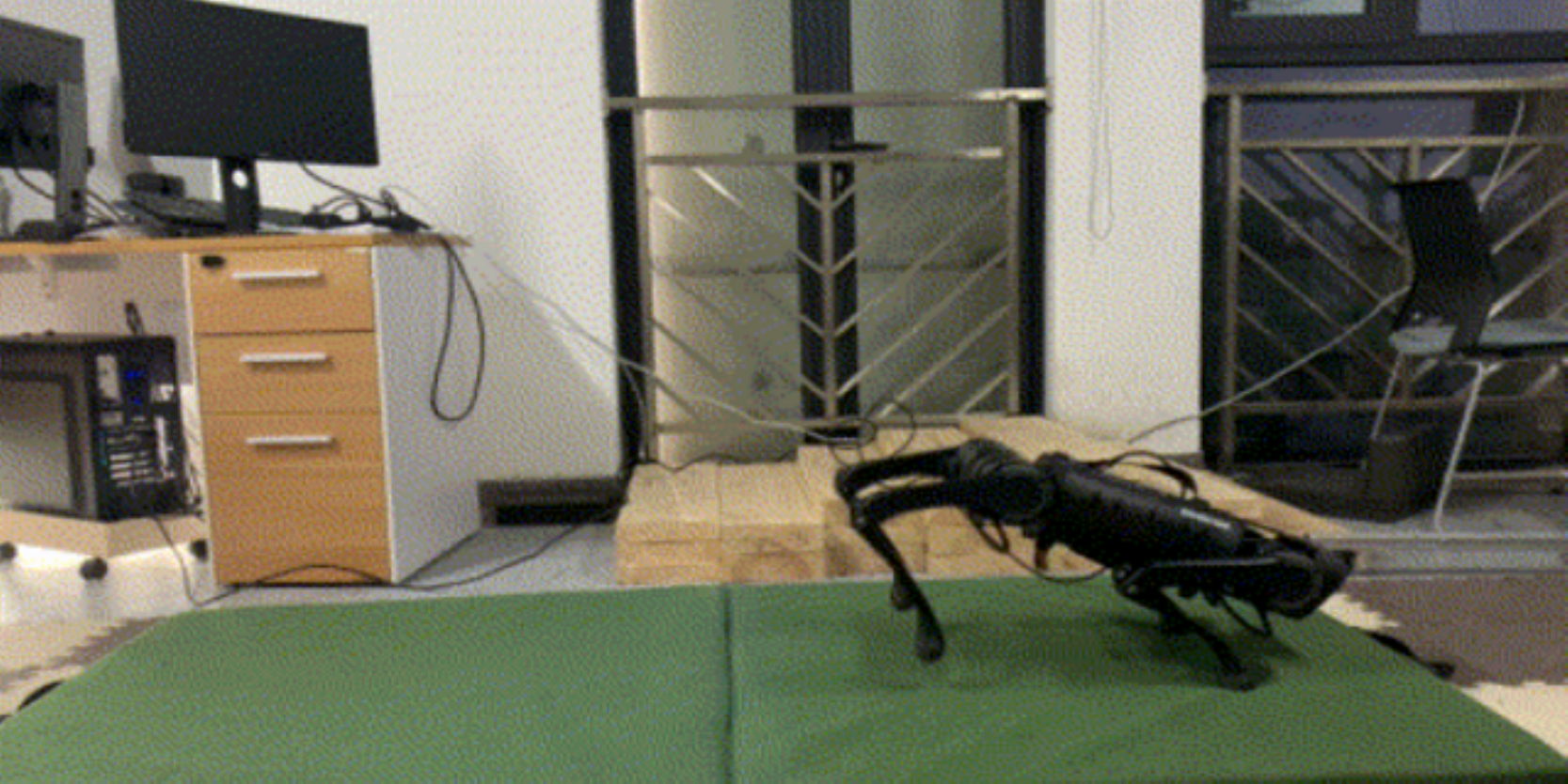}
    \includegraphics[width=0.11\textwidth]{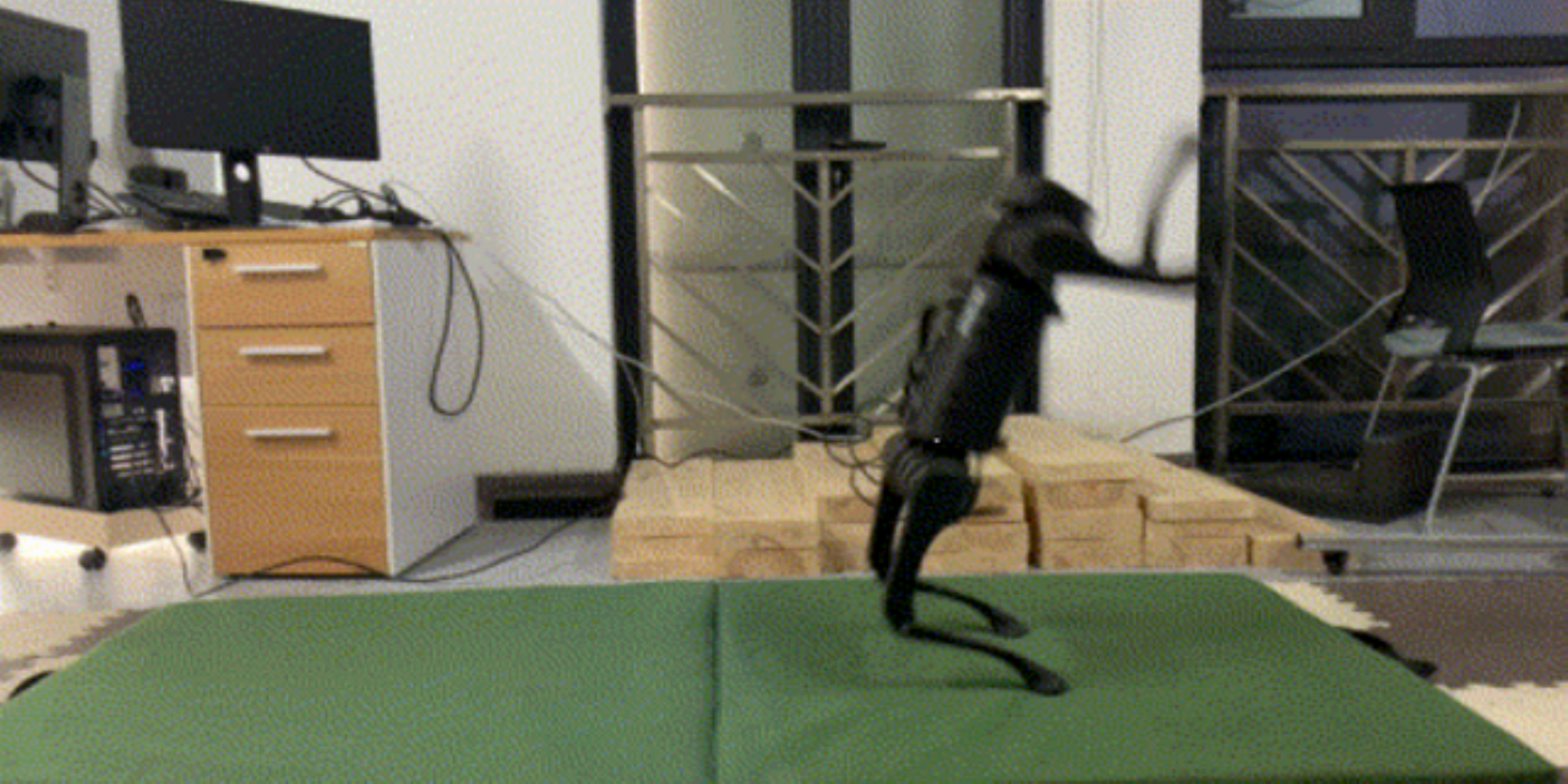}
    \includegraphics[width=0.11\textwidth]{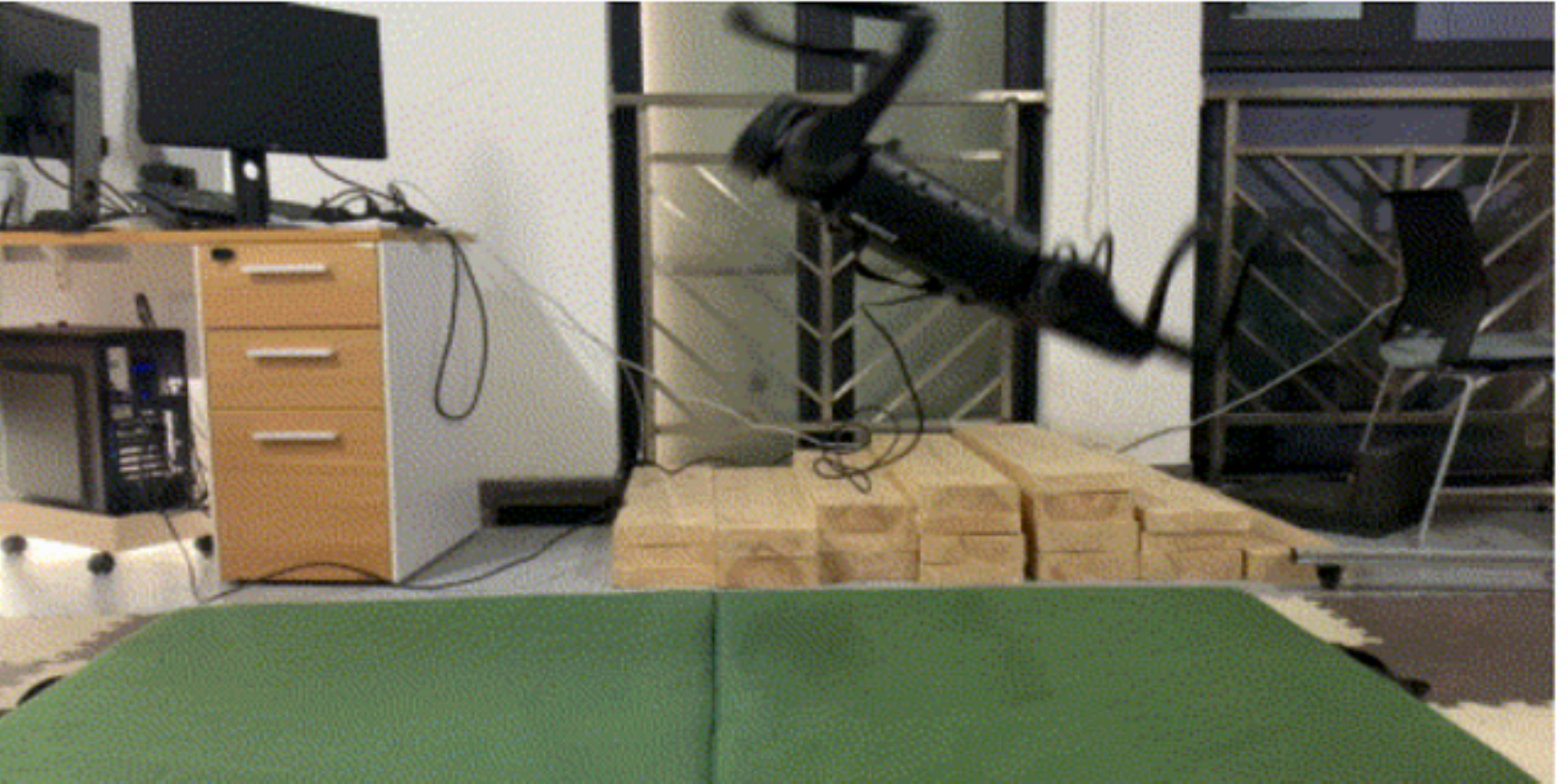}
    \includegraphics[width=0.11\textwidth]{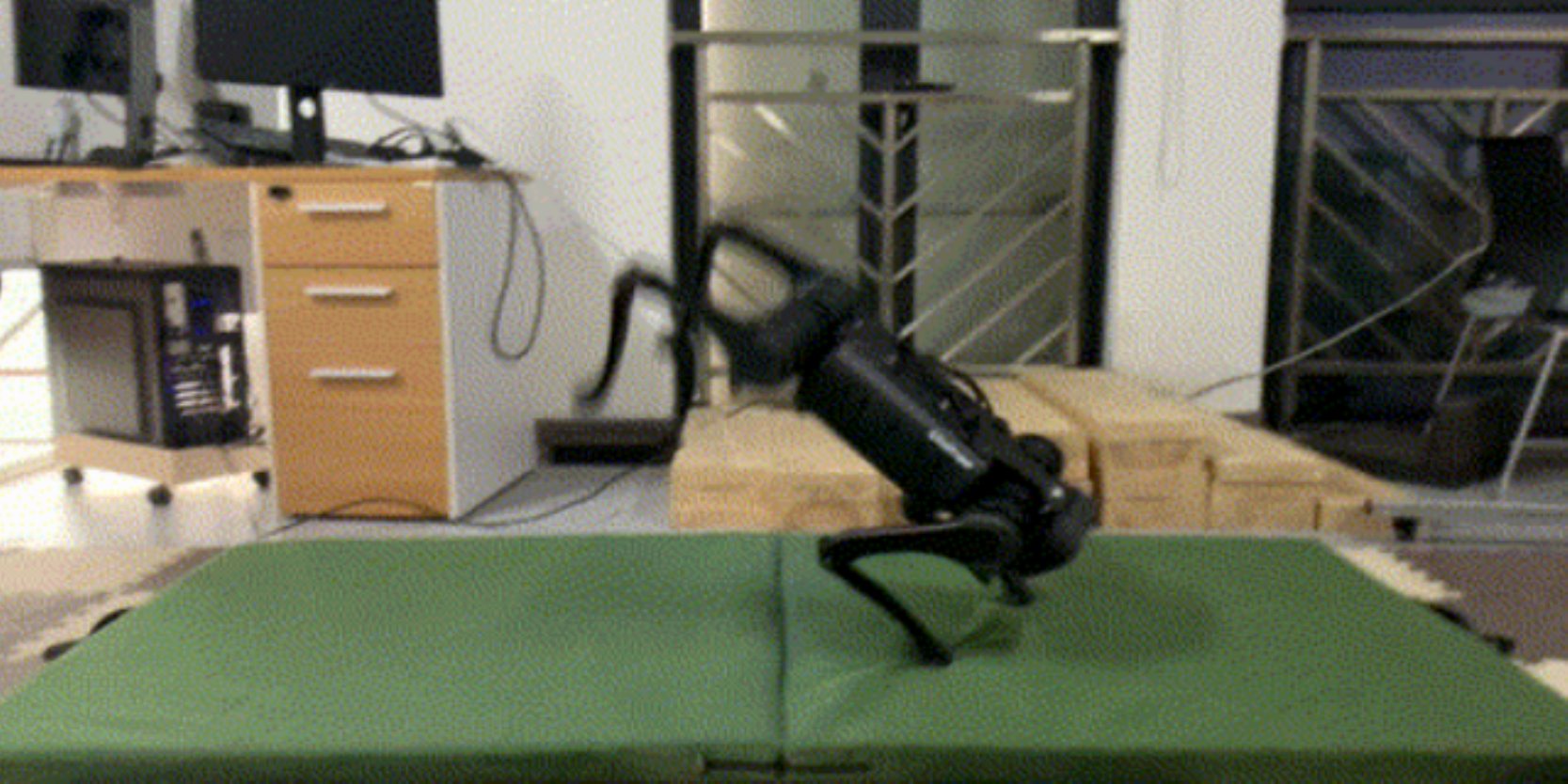}
    \includegraphics[width=0.11\textwidth,]{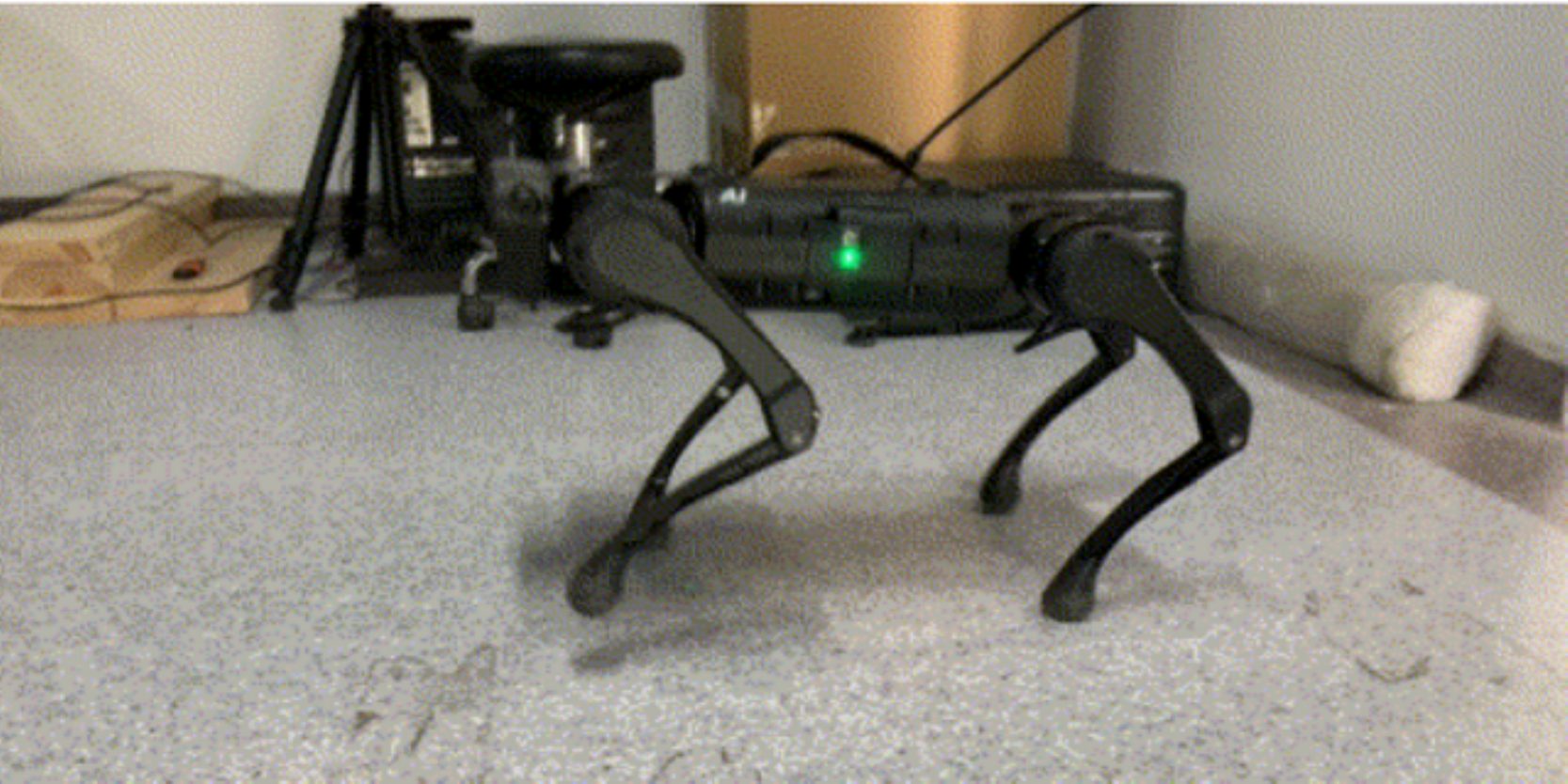}
    \includegraphics[width=0.11\textwidth]{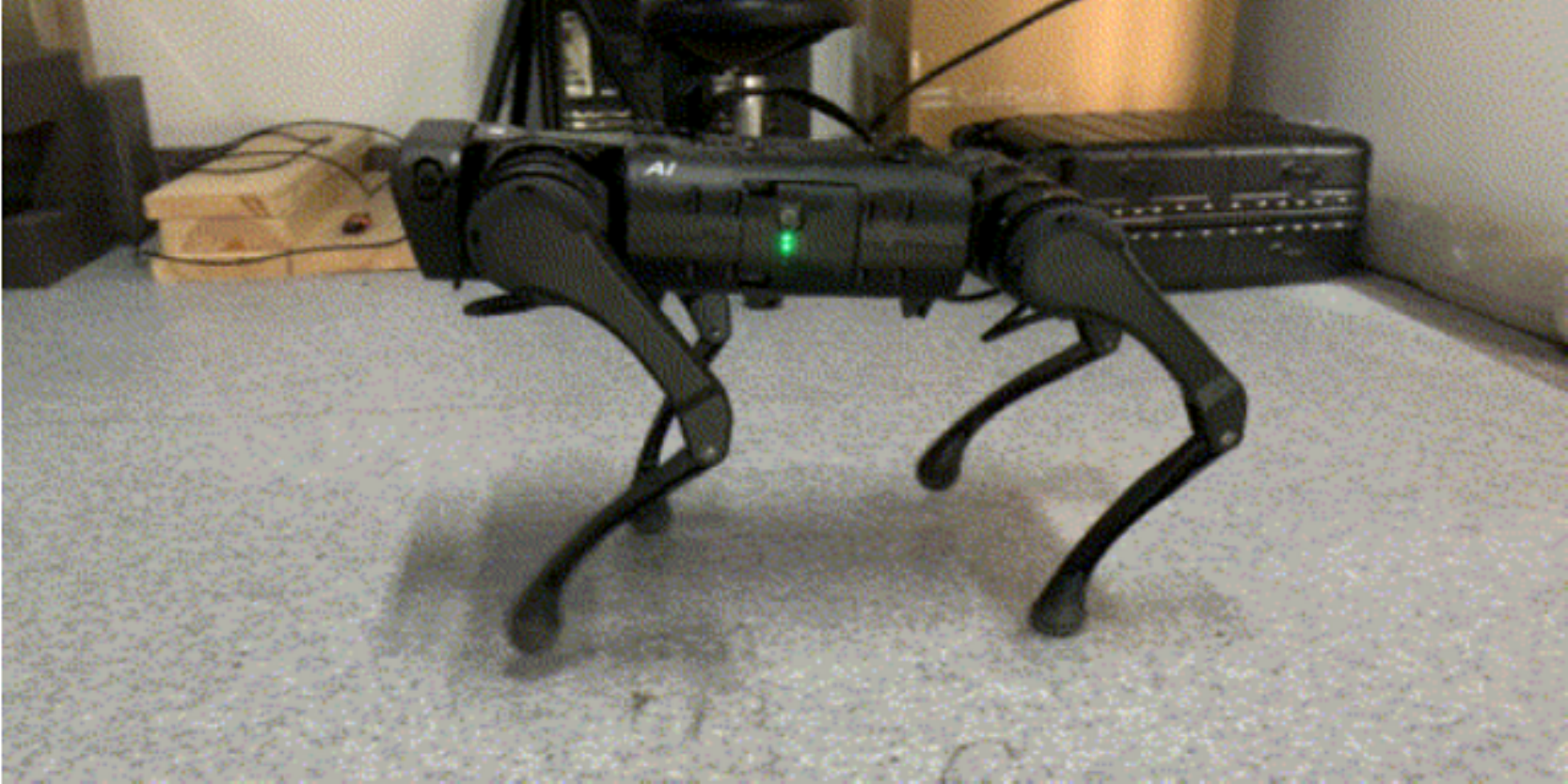}
    \includegraphics[width=0.11\textwidth]{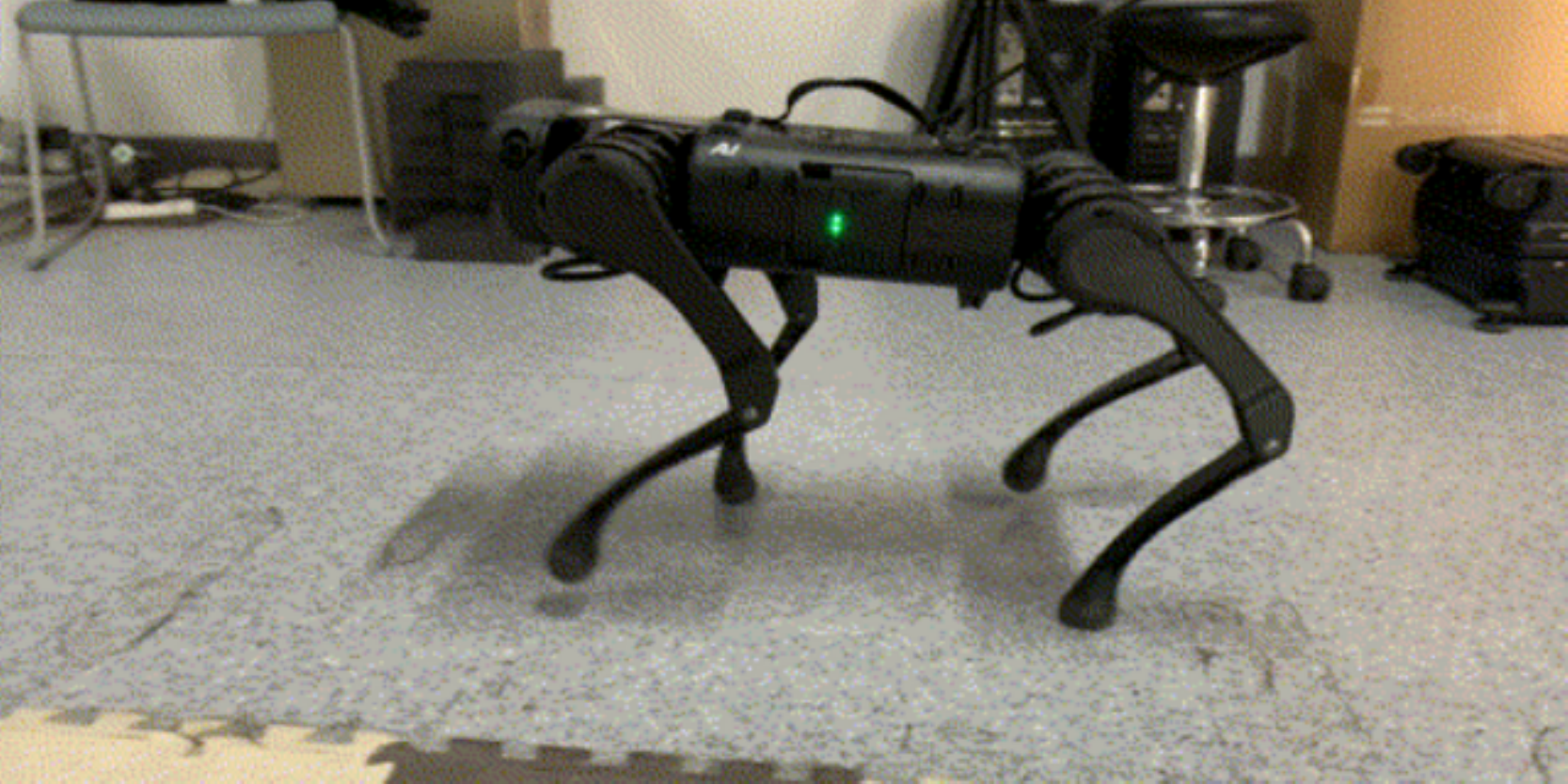}
    \includegraphics[width=0.11\textwidth]{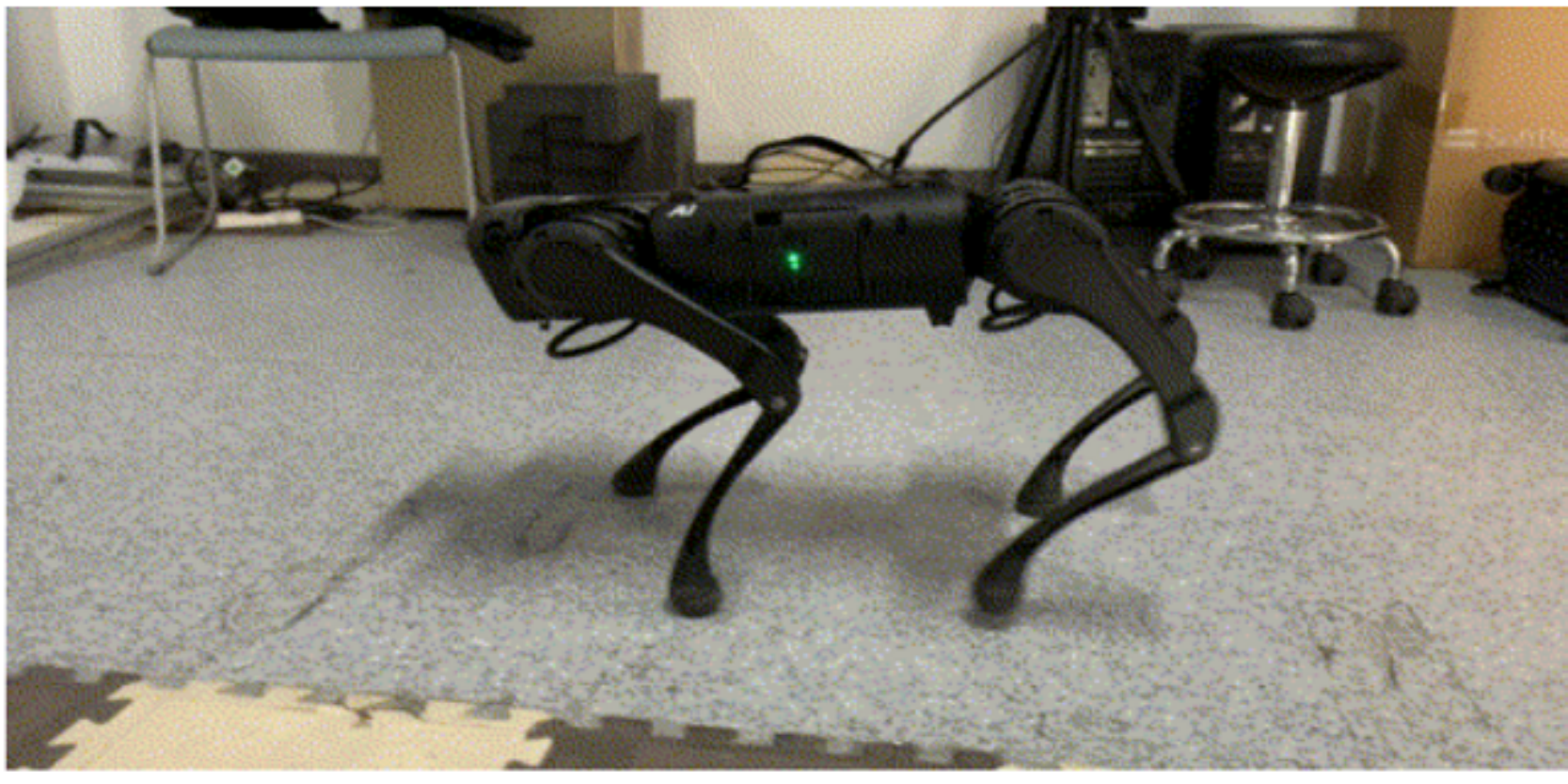}
    \caption{
    Real robot experiment of periodic and aperiodic motion.
    }
\label{Fig.flip_real}
\end{figure}

To solve this sim-to-real problem, we added domain randomization with randomized variables. We successfully transferred the skills learned from a walking horse video and backflip to the real robot as shown in Fig. \ref{Fig.flip_real}, which illustrates the stability of our model.

\section{CONCLUSIONS}
In this paper, we proposed an adaptive imitation framework that learns skills from a few seconds of animal movement video based on consistency and transfers the periodical gait and nonperiodic motion from different animals to the quadruped robot without any additional information. 

In the future, we will establish a high-level network that dynamically allows robots to learn new skills with few sets of data based on the existing skills and choose an optimal skill according to the tasks and learned skills.








\bibliographystyle{IEEEtran}
\bibliography{reference}

\end{document}